\documentclass[sigconf]{acmart}

\usepackage{enumitem}
\usepackage[ruled,vlined]{algorithm2e}
\usepackage{multirow}

\AtBeginDocument{%
  }

\copyrightyear{2024}
\acmYear{2024}
\setcopyright{rightsretained}
\acmConference[CODS-COMAD Dec '24]{8th International Conference on Data Science and Management of Data (12th ACM IKDD CODS and 30th COMAD)}{December 18--21, 2024}{Jodhpur, India}
\acmBooktitle{8th International Conference on Data Science and Management of Data (12th ACM IKDD CODS and 30th COMAD) (CODS-COMAD Dec '24), December 18--21, 2024, Jodhpur, India}
\acmPrice{}
\acmDOI{10.1145/3703323.3703337}
\acmISBN{979-8-4007-1124-4/24/12}

\begin{document}

\title{A Scalable Approach to Covariate and Concept Drift Management via Adaptive Data Segmentation}

\author{Vennela Yarabolu}
\affiliation{%
\institution{Computer Science \\ Indian Institute of Technology, Bombay}
\city{Mumbai}
\country{India}}
\email{210050168@iitb.ac.in}
\authornote{Work done during an internship at Mastercard, India}

\author{Govind Waghmare}
\affiliation{%
\institution{Mastercard}
\city{Gurugram}
\country{India}}
\email{govind.waghmare@mastercard.com}

\author{Sonia Gupta}
\affiliation{%
\institution{Mastercard}
\city{Gurugram}
\country{India}}
\email{sonia.gupta@mastercard.com}

\author{Siddhartha Asthana}
\affiliation{%
\institution{Mastercard}
\city{Gurugram}
\country{India}}
\email{siddhartha.asthana@mastercard.com}

\renewcommand{\shortauthors}{Yarabolu et al.}

\begin{abstract}
  In many real-world applications, continuous machine learning (ML) systems are crucial but prone to data drift—a phenomenon where discrepancies between historical training data and future test data lead to significant performance degradation and operational inefficiencies. Traditional drift adaptation methods typically update models using ensemble techniques, often discarding drifted historical data, and focus primarily on either covariate drift or concept drift. These methods face issues such as high resource demands, inability to manage all types of drifts effectively, and neglecting the valuable context that historical data can provide. We contend that explicitly incorporating drifted data into the model training process significantly enhances model accuracy and robustness. This paper introduces an advanced framework that integrates the strengths of data-centric approaches with adaptive management of both covariate and concept drift in a scalable and efficient manner. Our framework employs sophisticated data segmentation techniques to identify optimal data batches that accurately reflect test data patterns. These data batches are then utilized for training on test data, ensuring that the models remain relevant and accurate over time. By leveraging the advantages of both data segmentation and scalable drift management, our solution ensures robust model accuracy and operational efficiency in large-scale ML deployments. It also minimizes resource consumption and computational overhead by selecting and utilizing relevant data subsets, leading to significant cost savings. Experimental results on classification task on real-world and synthetic datasets show our approach improves model accuracy while reducing operational costs and latency. This practical solution overcomes inefficiencies in current methods, providing a robust, adaptable, and scalable approach to maintaining high-performance ML systems across various applications.

\end{abstract}





\maketitle

\section{Introduction}
In the rapidly evolving landscape of modern technology, ML systems have become indispensable for various applications. However, these systems are frequently challenged by data drift—a phenomenon where discrepancies between historical training data and future test data cause significant performance degradation and operational inefficiencies. Addressing data drift is critical to maintaining the reliability and efficiency of ML systems, yet traditional methods often fall short in several respects. Data drift is classified into two main categories: covariate shift and concept drift \cite{MORENOTORRES2012521,gama_2014_survey_drift,shimodaira2000improving,widmer1996}. Covariate shift is a type of data drift where the distribution of the input features (covariates) changes between the training and test datasets, but the relationship between the input features $X$ and the target variable $y$ (the conditional distribution $P(y \vert X)$) remains the same. Concept drift refers to a phenomenon where the relationship between the input data (features) and the target variable changes over time. This change affects the conditional distribution $P(y \vert X)$, meaning the way the output is generated from the input features evolves. Concept drift can significantly degrade model performance if not properly managed. Common methods like periodic retraining and re-weighting recent data are often ineffective, leading to accuracy drops and performance variations. Traditional approaches usually focus on either covariate or concept drift, often neglecting comprehensive solutions and discarding valuable historical data. This paper argues for incorporating drifted data into the training process to enhance model accuracy and robustness. We propose a scalable framework that combines data-centric approaches with adaptive management of both covariate and concept drift. Our solution uses sophisticated data segmentation to select optimal data batches for training, ensuring models remain accurate over time.

Our framework integrates data segmentation and drift management to enhance model accuracy and efficiency in large-scale ML deployments. By focusing on relevant data subsets, we reduce resource use, lowering costs and latency. Experimental results show improved accuracy, reduced costs, and adaptability to evolving data. The framework addresses both covariate shift and concept drift, maintaining model performance over time, and easily integrates with existing ML pipelines for smooth transitions and tracking. This approach enables organizations to maintain high-quality predictions and informed decisions in dynamic data environments.
\noindent The summary of our contributions is as follows:
\begin{itemize}
    \item \textbf{Scalability and Efficiency}: We introduce a robust framework designed to handle data drift, including both concept drifts and covariate shifts. Our approach is scalable and efficient, combining the strengths of data-centric methods with multiple drift management techniques.
    \item \textbf{Adaptive Data Subset Selection}: We develop an efficient data subset selection algorithm that initially identifies core data segments while discarding those affected by concept drift. Subsequently, it selects core data batches from these segments that are similar to the test set, thereby mitigating covariate shift. These steps reduce the amount of data required for training leading to operational efficiencies.
    \item \textbf{Optimal Performance}: Extensive experiments on synthetic and real datasets demonstrate that our method achieves better results while maintaining efficiency. Ablation on the trade-off between the $\%$ of data used and prediction accuracy underscore the cost benefits for practical deployments. 
\end{itemize}

\section{Related Work}
Drift detection is crucial in the environments where data distributions evolve. When drifts occur, the model performance can drop. Various drift detection techniques (\cite{lu2018learning,webb2016characterizing,Krawczyk2017EnsembleLF, gama_2014_survey_drift}) have been developed to identify drifts by pinpointing change points or time intervals \cite{basseville1993detection}. Effective drift detection methods ensure that models remain accurate and relevant by signaling the need for retraining or adjustments, thereby allowing the model to adapt to the new data distribution. These techniques are broadly classified into supervised methods, such as DDM \cite{gama2004learning}, EDDM \cite{baena2006early}, and ADWIN \cite{bifet2007learning}, and unsupervised methods, like HDDDM \cite{ditzler2011hellinger} and DAWIDD \cite{hinder2020towards}. 

Model-centric and data-centric approaches address data drift differently. Model-centric methods, like retraining and online learning, adapt models to changing patterns, enhancing adaptability but at high cost and complexity \cite{gomes_2017,Elwell_2011,song_2021,bifet_2009,domingos_2000,gama_2003}. Data-centric strategies, such as subset selection and reweighting, ensure training data remains relevant, improving efficiency. Combining both manages drift effectively, balancing accuracy and resource use to maintain model reliability and performance in dynamic environments. This hybrid approach ensures models stay robust against evolving data trends over time.

In addition to model-centric methods, data-centric approaches have been developed to adapt to concept drift. Data reduction techniques \cite{ramirez_2017} focus on cleaning data by removing noisy samples and features. Drift understanding techniques \cite{dong_2021_Cybernetics} filter out obsolete data using the newest data segment as a pattern, based on cumulative distribution function comparisons. Once filtered out, samples are not reselected, even if they could be beneficial later. A notable technique in this category is CVDTE \cite{fan_2004}, which aims to select samples that do not yield conflicting predictions between previous and current models. However, these techniques share a fundamental limitation: they lack mechanisms to validate if the data preprocessing steps genuinely enhance model accuracy. In comparison, we take a more data-driven approach by explicitly evaluating models on selected data segments, while minimizing computational costs.

The issue of data drift, which refers to the variation in model accuracy over time, has been a significant area of research in the ML community. It arises when the model encounters test data that differ substantially from the training data, leading to reduced prediction accuracy. Numerous studies have explored this challenge, particularly in the context of streaming or continuously arriving data, and have proposed various approaches to address it \cite{bifet_2007, MORENOTORRES2012521, tahmasbi2020driftsurfriskcompetitivelearningalgorithm, suprem2020odin}. In supervised learning tasks, where features $X$ are used to predict labels $y$, data drift is commonly caused by two factors \cite{MORENOTORRES2012521}: (1) Covariate shift, which occurs when the distribution of features $X$ changes, such as when new types of incidents with previously unseen feature values arise; and (2) Concept drift, which happens when the underlying relationship between features $X$ and labels $y$ shifts, for example, due to changes in a system and its dependencies, leading to different causal relationships between symptoms and components. To tackle the problem of data drift, existing methods can be categorized into three main types: (a) Window-based approaches \cite{widmer1996}, which employ a sliding window of recent data for training updated models; (b) Shift detection methods \cite{pesaranghader2018mcdiarmid}, which utilize statistical tests to identify the occurrence of data drift and trigger model retraining only when such shifts are detected; and (c) Ensemble-based strategies \cite{Brzezinski_2014}, which create ensembles of models trained on previous data, combining their predictions through a weighted average to maintain accuracy.

\section{Background}

\subsection{Covariate Shift}

Covariate shift occurs when the distribution of input features changes between training and test datasets, while their relationship with the target variable remains the same \cite{shimodaira2000improving,kelly1999impact}. This shift can result from environmental changes, data collection methods, or sampling procedures. Addressing covariate shift involves reweighting training data or using domain adaptation methods to maintain model accuracy on new data. Most approaches focus on training and test datasets without considering continuous time \cite{ditzler2015learning}, addressing the issue by modifying training objectives or adjusting the importance of training data to improve test accuracy \cite{shimodaira2000improving,sugiyama2007covariate,yamazaki2007asymptotic,mansour2008domain,gretton2008covariate}. For input features $X$, the covariate shift is defined as follows:

\begin{equation}
    P_{Train}(X) \neq P_{Test}(X)
\end{equation}

\subsection{Concept Drift} 

Concept drift occurs when the relationship between input features and the target variable evolves over time \cite{tsymbal2004problem}. This shift can happen gradually or suddenly, altering the underlying data patterns. Concept drift challenges ML models by potentially decreasing their accuracy and reliability if not detected and addressed. To handle concept drift effectively, it's essential to continuously monitor model performance and update the model to adapt to new data patterns, ensuring sustained accuracy and relevance. Broadly, there are three solution categories to handle concept drift as window-based \cite{widmer1996, Koychev_2006, Klinkenberg_2004}, detection-methods \cite{gama2004learning,kifer_2004,bifet_2007,Pesaranghader_2016,pesaranghader2018mcdiarmid,tahmasbi2020driftsurfriskcompetitivelearningalgorithm} and ensemble methods \cite{Elwell_2011,Brzezinski_2014,sun2018concept,zhao_2020}. For input features $X$ and target  $y$, the concept drift is defined as follows:

\begin{equation}
    P_{Train}(y \vert X) \neq P_{Test}( y \vert X)
\end{equation}

\section{Overview}

\begin{figure*}[h]
  \includegraphics[width=\textwidth]{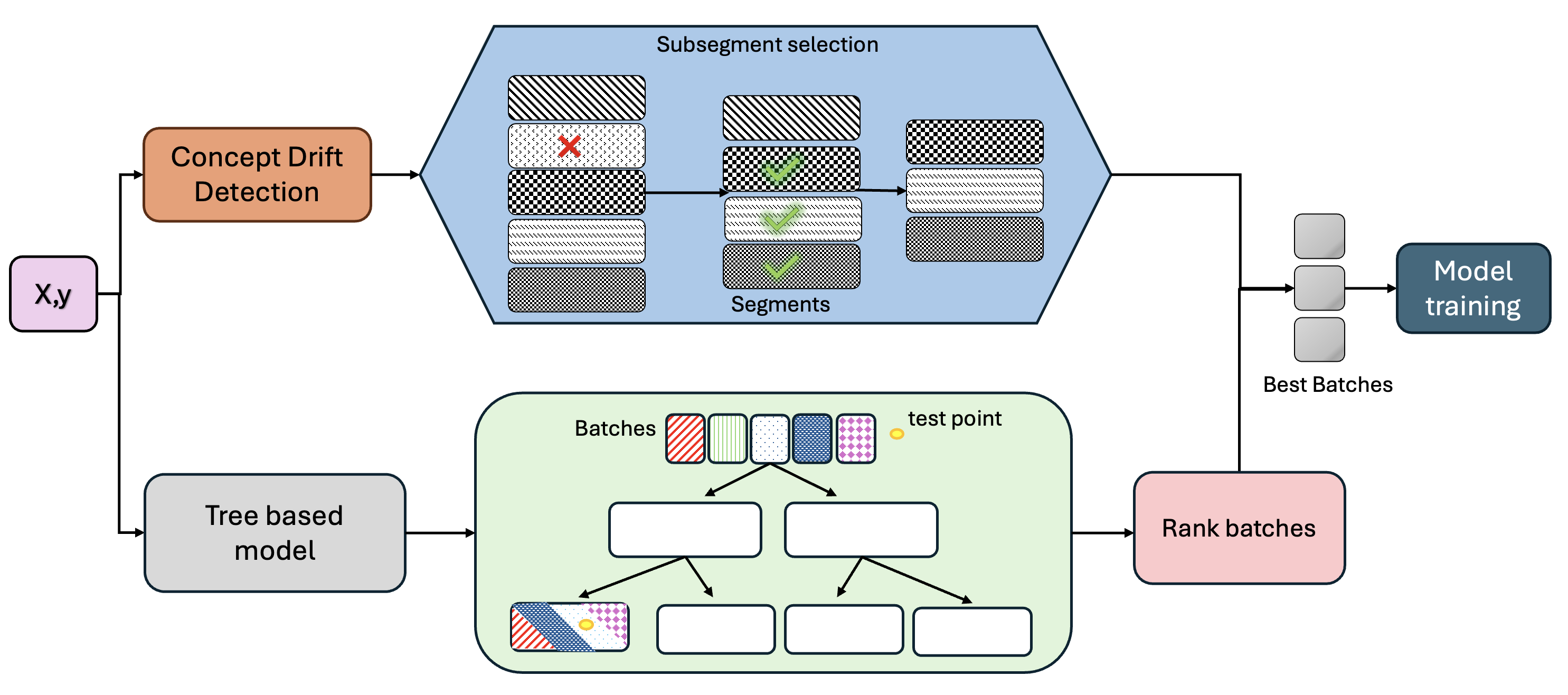}
  \caption{ Complete architecture of our method. The top branch focuses on concept drift while bottom branch ranks train batches based on covariate shift.}
  \Description{}
  \label{fig:main_arch}
\end{figure*}

The overall process of our method is shown in Figure \ref{fig:main_arch}. Our approach tackles the two primary causes of data drift: covariate shift and concept drift. The idea is to select the most relevant batches from the training data segments based on their relationship to the test samples, and use those batches to train the model for accurate inference. Inline with \cite{MLSYS2022_matchmaker}, this method is based on two conjectures: 

\noindent \textbf{(1)} If the decline in accuracy is due to the training and test data residing in different regions of the data space (covariate shift), it is logical to prioritize the training batch $t$ whose features ($X_t$) are closest to those of the test data ($x_{*}$). 

\noindent \textbf{(2)} If the accuracy drop results from changes in the \( x \rightarrow y \) relationship over time (concept drift), then it is prudent to exclude data segments that exhibit concept drift relative to the current data segment, as indicated by gradient scores.

To address covariate shift, a random forest $R$ is trained meticulously on all labeled training batches $\{(X_1, y_1), \ldots, (X_T, y_T)\}$. This sophisticated technique partitions the training data space, harnessing the random forest's prowess in organizing complex data distributions \cite{davies2014random,MLSYS2022_matchmaker}. During testing, we rank the training batches based on their similarity to the test sample by analyzing the leaf nodes in $R$ where the test sample is mapped. Batches are then ranked according to the concentration of training points that fall within these leaf nodes, ensuring that the most pertinent data is utilized to refine model accuracy. This process is highlighted in the Figure \ref{fig:covariate_shift}.

We build on top of concept drift approach Quilt \cite{kim2024quilt}.
The concept drift detection component monitors each sample in the data stream to identify potential changes in data patterns. Various drift detection methods can be employed, based on shifts in data distribution or model performance. When a drift is detected, a new data segment is created from the drift point and becomes the current segment. The data segment selection component then updates the model using selected segments. If no drift is detected, the sample is added to the existing segment. The framework performs two main operations for selecting data segments: (1) discarding segments that no longer align with the current data pattern, and (2) selecting a core subset of stable segments for efficient model training. This approach leverages gradient-based disparity and gain scores as described in the \cite{kim2024quilt}, which are computationally efficient and independent of specific data characteristics, unlike traditional statistical distance measures that can struggle with high-dimensional data and scalability issues. This method allows for adaptive handling of data drift without needing ground truth labels for retraining.

\section{Data Selection }

\subsection{Covariate Shift Ranking}

To prioritize training data segments based on covariate shift, we focus on their proximity to test points in the data space. Although one might consider ranking training batches by their average Euclidean distance from the test point, this method has limitations. Euclidean distance computation becomes expensive with larger batches, is prone to outliers \cite{fischler_1981}, and struggles with high-dimensional data. Instead, we recommend using decision trees and random forests for ranking batches, similar to \cite{MLSYS2022_matchmaker}. This approach scales well, is more robust to outliers, and handles high-dimensional data more effectively, making it a practical choice for complex datasets. We rely on methods described in MatchMaker \cite{MLSYS2022_matchmaker} to rank batches using decision trees and random forest as follows:

\paragraph{Covariate Shift Ranking of the Batches}

Decision trees classify data by partitioning it at feature thresholds that optimize prediction accuracy, grouping similar samples into the same leaf nodes. When a new sample is tested, it is routed to a leaf node, and its label is predicted based on the majority label within that node. We use this mechanism to evaluate training batches for covariate shift, prioritizing those that are closer to the test sample. Let $\{(X_1, y_1), \ldots, (X_T, y_T)\}$ denote training batches. Once decision tree is constructed on these batches, let $N[k][t]$ indicate the number of samples from batch $t$ that fall into leaf node $k$. Now, for a test point that is assigned to leaf node $k_{*}$, one has to calculate covariate shift ranking $\textbf{Rank}_{cov\_shift}$ of the training batches. This ranking is computed by ordering $N[k_{*}][t]$ starting from lowest covariate shift to highest as shown:

\begin{equation}
    \textbf{Rank}_{cov\_shift} = \text{argsort}\{N[k_{*}][1], \ldots, N[k_{*}][T]\}
\end{equation}

\begin{figure}[t]
  \centering
  \includegraphics[width=\linewidth]{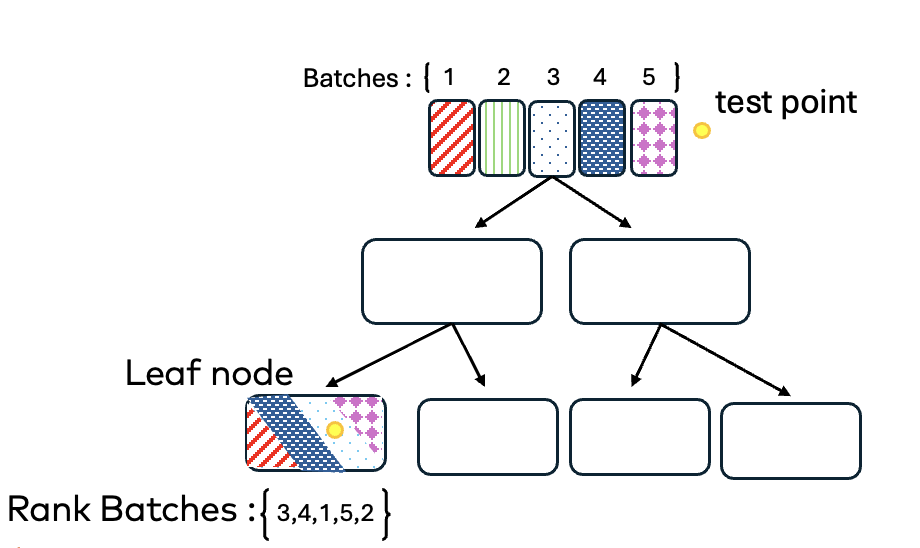}
  \caption{Covariate shift ranking of training batches 1,2,3,4 and 5 with respect to test point shown in yellow. In the leaf node, batches are ranked as 3,4,1,5 and 2. Here, batch 3 has lowest covariate shift.}
  \label{fig:covariate_shift}
\end{figure}

\noindent As random forest are capable of modeling high-dimensional data, this approach is extended to the random forests for better performance. The visual summary of covariate-ranking of the batches is shown in the Figure \ref{fig:covariate_shift}. 

\subsection{Concept Drift}

To tackle the concept drift, our approach employs a robust framework Quilt \cite{kim2024quilt}, centered on two key tasks: (1) removing data segments that show concept drift compared to the current segment, and (2) selecting a core set of stable data segments to train the model efficiently while maintaining accuracy. This method utilizes disparity and gain scores, calculated from gradient values on training and validation sets, ensuring minimal computational cost. In the next sections, we discuss the gradient computation along with disparity and gain score formulation as defined in the Quilt \cite{kim2024quilt}.

\noindent {\textbf{Gradient Computation}}: The last layer of the neural network calculates the logits for each class. Let $X_i' \in \mathbb{R}^{d'}$ be the embedding feature of the $i$th input data $X_i$ with a hidden layer dimension of $d'$, and $z_i \in \mathbb{R}^c$ be the logit outputs computed by $z_i = w \cdot X_i' + b$ using the last layer weights $w \in \mathbb{R}^{d' \times c}$ and bias $b \in \mathbb{R}^c$. To convert a logit $z_i$ into a probability vector $\hat{y}_i$, softmax function is used as follows: $ \hat{y}_i = \text{softmax}(z_i) = \frac{e^{z_{ij}}}{\sum_{j=1}^c e^{z_{ij}}}.$
We can also rewrite the model output $\hat{y}_i$ as a function of the model parameters $\theta$ and input data $X_i$ as $\hat{y}_i = f_\theta(X_i)$. Given the model output $\hat{y}_i$ and the true label $y_i$, the cross-entropy loss between them is $L_i = L(y_i, \hat{y}_i) = - \sum_{j=1}^c y_{ij} \log(\hat{y}_{ij})$. The last layer gradient approximation is given as $g = (\nabla_b L, \nabla_w L)$ where gradients of the front layers are not used. Using the chain rule, on can compute the gradient of the $i^{\text{th}}$ sample as follows:
$g = \left( \nabla_b L, \nabla_w L \right) = \left( \hat{y}_i - y_i, (\hat{y}_i - y_i) \cdot X_i' \right).$

\noindent {\textbf{Disparity Score}}: The disparity score ($D$), is a measurement of dissimilarity between two data distributions. It detects segments exhibiting concept drift. Concept drift is characterized by a change in the posterior distribution $P(y|X)$ while the data distribution $P(X)$ remains constant. Essentially, it reflects variations in the predicted labels $y$ for the same input data. To quantify this change, one can use the measure $\mathbb{E}[\|y_t - y_v\|]$, which represents the expected label difference between a training subset and a validation set, where $y_t$ and $y_v$ denote the true labels from the training and validation sets, respectively. This measure is analogous to the concept drift severity \cite{minku2009impact}. Direct computation of this measure is computationally expensive as it requires identifying similar samples across the training and validation sets and comparing their label differences. To overcome this, Quilt \cite{kim2024quilt} proposed a gradient-based score as an efficient approximation as follows. The disparity score $D$ of a training subset $T$ with respect to a validation set $V$ is defined as:
\begin{equation}
D(T,V) = \left\| \frac{1}{|T|} \sum_{t=1}^{|T|} g_t - \frac{1}{|V|} \sum_{v=1}^{|V|} g_v \right\| = \left\| \mathbb{E}[g_t] - \mathbb{E}[g_v] \right\|,
\end{equation}

\noindent where $|V|$ denotes the size of the validation set. Also, the $D$ score measures the $L2$-norm distance between two gradient vectors. 

\begin{figure}[t]
  \centering
  \includegraphics[width=\linewidth]{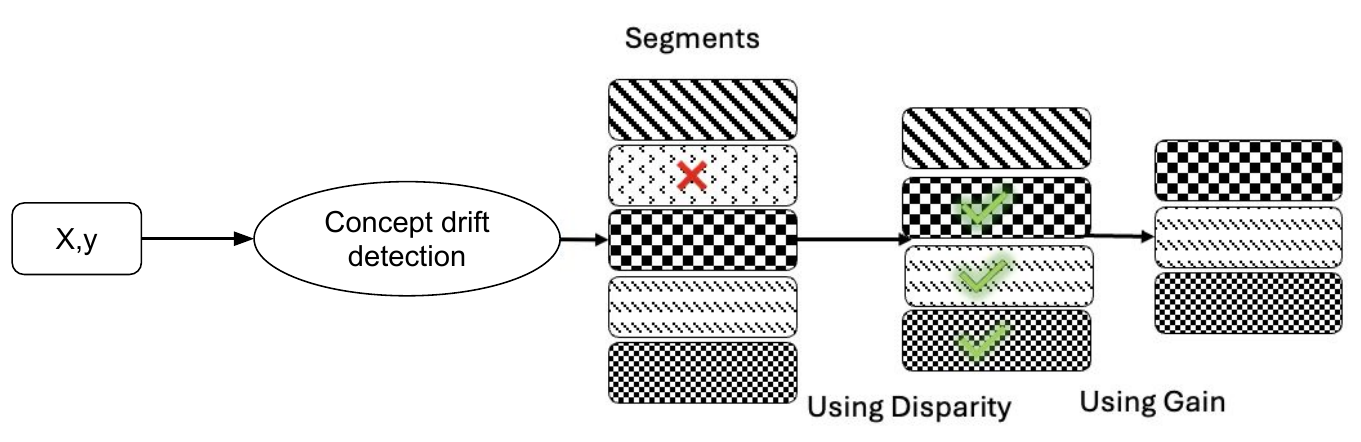}
  \caption{Subsegment selection approach using gradient based disparity and gain scores}
  \label{fig:subsegment_selection}
\end{figure}

\noindent {\textbf{Gain Score}}: The gain score is built on top of methods introduced in \cite{killamsetty2021glistergeneralizationbaseddata, killamsetty2021gradmatchgradientmatchingbased}. Consider historical data for both training and validation. Studies indicate that selecting a subset where the inner product of the average gradients between the subset and the validation set (known as the gain) is positive can lower the model’s validation loss during training. Essentially, gradient vectors represent the direction and size of updates in gradient descent, and aligning these gradients between the training and validation sets helps improve model performance. The gain score $G$ for a training subset $T$ with respect to a validation set $V$ is:
\begin{equation}
G(T,V) = \frac{1}{|T|} \sum_{t=1}^{|T|} g_t \cdot \frac{1}{|V|} \sum_{v=1}^{|V|} g_v = \mathbb{E}[g_t] \cdot \mathbb{E}[g_v],
\end{equation}
\noindent where $\cdot$ represents the dot product of the gradient vectors. The subsegment selection procedure based on gradient based disparity and gain score is highlighted in the Figure \ref{fig:subsegment_selection}.

\section{Data Subset Selection Algorithm}

Algorithm 1 provides a method for detecting covariate shift by examining how sample distributions vary across different batches through the lens of a trained model's decision trees. Initially, the entire training data segments are utilized to train the model \(R\). For each individual decision tree \(T_i\) within the model, the algorithm computes a score \(S[k][t]\) for every batch \(t\) across each leaf node \(k\). This score quantifies how many other batches within the same leaf node contain fewer samples \(N[k][t]\) compared to the batch \(t\) under consideration. By evaluating these scores, the algorithm can detect shifts in feature distributions among different batches, which may signal potential covariate shifts.

Algorithm 2 outlines the procedure for selecting data segments to optimize model training. The process starts by initializing the model parameters and proceeds through a series of epochs. For each epoch, the algorithm initializes an empty training subset \( S \). It then calculates the average gradient over the validation set \( d_{V_N} \). Next, the algorithm iterates over previous data segments to compute their gradient averages and evaluates their gain and disparity scores. Data segments with a positive gain score and a disparity score below a specified threshold are added to the training subset \( S \). The current training da \( d_{T_N} \) is always included in \( S \) to ensure recent data is used in training.

Additionally, the algorithm initializes an empty set for the best batches \( B_{\text{best}} \). For each sample \( v \) in the validation set \( d_{V_N} \), it retrieves the rankings of the batches that we had from algorithm 1 based on the mapped leaf of \( v \) in a random forest (rf). The algorithm then iterates through these ranked batches and adds them to \( B_{\text{best}} \) if they are part of the selected segments \( S \), breaking the loop once a suitable batch is found. This ensures that the best batches from the validation set, which are also part of the selected training segments, are prioritized.The model parameters are updated using the learning rate \( \eta \) and the computed gradients from the best batches \( B_{\text{best}} \). This process is repeated for \( T \) epochs. Finally, the algorithm returns the updated model parameters \( \theta_T \).

\begin{algorithm}[h]
\caption{Covariate Shift Scoring}
\label{alg:covariate}
    \KwIn{ Training data batches$\{(X_1, y_1), \ldots, (X_T, y_T)\}$}
    \KwOut{Stored values $S[k][t]$ for each tree $T$ }
    Train model $R$ on entire data $(X_1, y_1), \ldots, (X_T, y_T)$\;
\For{each tree $T_i \in R$}{
    Store $S^{i}[k^{i}][t] = \sum_{(t \neq t')} \{  1 \quad \text{if } N[k^{i}][t] > N[k^{i}][t'] \}$\;
}
\end{algorithm}

\begin{algorithm}
\caption{Data Selection Algorithm }
\label{alg:concept}
\KwIn{Previous data segments $D_{\text{prev}} = \{d_1, \ldots, d_{N-1}\}$, current segment $d_{T_N}$, validation set $d_{V_N}$, loss function $L$, learning rate $\eta$, maximum epochs $T$, disparity threshold $T_d$, batch size $B$, number of estimators $n_{\text{estimators}}$, maximum depth $d_{\text{max}}$}
\KwOut{Final model parameters $\theta_T$}
\For{epoch $t$ in $[1, \ldots, T]$}{
    Initialize training subset $S = \emptyset$\;
    $g_V = \frac{1}{|d_{V_N}|} \sum_{j=1}^{|d_{V_N}|} g_j$\;
    \For{segment $d$ in $D_{\text{prev}}$}{
        $g_d = \frac{1}{|d|} \sum_{k=1}^{|d|} g_k$\;
        $G_d = g_d \cdot g_V$\;
        $D_d = \|g_d - g_V\|$\;
        \If{$G_d > 0$ \textbf{and} $D_d < T_d$}{
            $S = S \cup d$\;
        }
    }
    $S = S \cup d_{T_N}$\;

    Initialize best batches $B_{\text{best}} = \emptyset$\;
    \For{each sample $v$ in $d_{V_N}$}{
        Get the rankings of the batches based on the mapped leaf of $v$ in rf\;
        \For{each batch in the rankings}{
            \If{batch is in $S$}{
                $B_{\text{best}} = B_{\text{best}} \cup \text{batch}$\;
                break \;
            }
        }
    }

    Update $\theta_t = \theta_{t-1} - \eta \frac{1}{ B_{\text{best}}} \sum_{e \in  B_{\text{best}}} \nabla_{\theta} L(e)$\;
}
\Return final model parameters $\theta_T$
\end{algorithm}

\section{Experiments}

\subsection{Datasets}

We conducted experiments using a varied selection of datasets, including five synthetic and five real-world datasets. Table \ref{tab:dataset_statistics} provides detailed descriptions and summary statistics for each dataset used in our research. The synthetic datasets were deliberately crafted to represent different forms of concept drifts, as outlined by \cite{lu2018learning}. All datasets expect Covcon are taken and preprocessed inline with \cite{kim2024quilt} while Covcon is taken and preprocessed as done in \cite{MLSYS2022_matchmaker}. 

\begin{itemize}[leftmargin=*]
    \item \textbf{SEA}: Streaming Ensemble Algorithm (SEA) is a standard dataset for simulating sudden concept drifts. The samples are in a 3D feature space with random numeric values between 0 and 10.
    \item \textbf{RandomRBF}: Random Radial Basis Function is used to make a number of random centroids and new samples are generated by selecting the center of centroids. 
     \item \textbf{Sine}: It contains four numerical features with values that range from 0 to 1. Two of the features are relevant to a given binary classification task, while the two other features simulate noise. 
     \item \textbf{Hyperplane}: Here, hyperplanes are viewed as concepts and varied orientiations are used to simulate drifts. A hyperplane is defined by feature weights, and weights are drifted over time. There are ten relevant features including two drift features.
    \item \textbf{Covcon and Covcon\_5M}: A 2-dimensional dataset to have covariate shift and concept drift. The decision boundary at each point is given by $\alpha * \sin(\pi x_1) > x_2$.
   
    
    \item \textbf{Electricity \cite{zliobaite2013good} :} This is Australian New South Wales Electricity Market data from 1996 to 1998, measured every 30 minutes.
     \item \textbf{Weather :} Points measures the weather in Bellevue NE during the period of 1949–1999. 
    \item \textbf{Spam :} It is consists of email messages from the Spam Assassin Collection. There are 9,324 samples of messages and a message is represented by 499 features of boolean bag-of-words. The labels denote whether a message is spam or not. 
    \item \textbf{Usenet1 \& 2 :} Two real datasets are based on the 20 news group collection with three topics: medicine, space, and baseball. Each sample contains messages about different topics, and a user labels them sequentially by personal interests whether the topic of a message is interesting (1) or junk (0). 
    \item \textbf{Covertype \cite{MLSYS2022_matchmaker}:} This dataset contains 581K samples describing 7 forest cover types for 4 region in the Roosevelt National Forest.

\end{itemize}

\begin{table*}
  \caption{Dataset statistics.}
  \label{tab:dataset_statistics}
  \begin{tabular}{lccccccccc}
    \toprule
    \textbf{Type} & \textbf{Dataset} & \textbf{Size} & \textbf{Features} & \textbf{Classes} & \textbf{Num. Segments} & \textbf{Segment size} & \begin{tabular}{@{}c@{}} \textbf{Num. batches} \\ \textbf{per segment} \end{tabular} & \textbf{Batch size} \\
    \midrule
    \multirow{5}{*}{Synthetic} & SEA & 16K & 3 & 2 & 8  & 2K & 20 & 100 \\
        & Random RBF & 16K & 10 & 2 & 8  & 2K & 20 & 100 \\
        & Sine & 16K & 4 & 2 & 8  & 2K & 20 & 100 \\
        & Hyperplane & 16K & 10 & 2 & 8  & 2K & 20 & 100 \\
        & Covcon & 10K & 2 & 2 & 5  & 2K & 2 & 1K \\
        & Covcon\_5M & 5M & 2 & 2 & 10  & 500K & 10 & 50K \\
        \midrule
        \multirow{5}{*}{Real} & Electricity & 43.2K & 6 & 2 & 10  & 4.32K & 20 & 216 \\
        & Weather & 18K & 8 & 2 & 10  & 1.8K & 20 & 90 \\
        & Spam  & 9.3K & 499 & 2 & 9  & 1.036K & 14 & 74 \\
        & Usenet1  & 1.5K & 99 & 2 & 5  & 300 & 2 & 150 \\
        & Usenet2  & 1.5K & 99 & 2 & 5  & 300 & 3 & 100 \\
        & Covertype  & 581K & 54 & 7 & 10  & 58.1K & 10 & 5.81K \\
    \bottomrule
  \end{tabular}
\end{table*}

\subsection{Model Training and Hyperparameters}
Our approach involves employing two main strategies for training our models. First, we utilize a Random Forest to partition (Algorithm \ref{alg:covariate}) and rank batches of the data segment based on a specified batch size. It addresses covariate shift. Then, we employ a simple neural network classifier trained using cross-entropy loss to deal with concept drift (Algorithm \ref{alg:concept}). For a random forest, we use grid search over batch size [grid over 3-5 values], number of estimators $n_{\text{estimators}}$, maximum depth $d_{\text{max}}$. For all experiments, $n_{\text{estimators}}=50$ and maximum depth $d_{\text{max}}=20$. Batch size is reported in Table \ref{tab:dataset_statistics}. In Algorithm \ref{alg:concept}, a neural network classifier with a single hidden layer with 256 nodes is employed. The value of disparity threshold for each data segment is calculated using Bayesian optimization with the search interval interval in (0,2) inline with \cite{kim2024quilt}. The learning rate is set to \(1 \times 10^{-3}\) and early stopping with patience 10 is used for termination with maximum number of epochs limited to 2000. For computation, we have used RTX Quadro with 24 GB of VRAM and 32 GB of RAM on Linux machine. Codebase is developed using PyTorch. The subset selection method discards the segments based on gain and disparity scores. Further, only relevant batches from the remaining segments are selected leading to reduction in data used for training. This value for each dataset is reported in Table \ref{tab:first} and \ref{tab:second} on the last line '\textit{\% of data used}'.

\subsection{Baseline Algorithms}

\begin{itemize}[leftmargin=*]
\setlength{\itemsep}{1pt} 
    \item \textbf{Na\"ive Methods:}
    \begin{itemize}[leftmargin=*]
        \item \textbf{Full Data and Current Segment:} Full data uses all data segments for training, ensuring maximum information but ignoring the relevance of older data. Current segment trains only on the latest data, assuming it to be the most relevant. 
    \end{itemize}
    \item \textbf{Model-centric Methods:}
    \begin{itemize}[leftmargin=*]
        \item \textbf{HAT} \cite{bifet2007learning}: Trains a Hoeffding Adaptive Tree classifier online using the full dataset, adapting to data distribution changes by incrementally updating its structure.
        \item \textbf{ARF} \cite{gomes_2017}: Implements an Adaptive Random Forest classifier on each sample, leveraging the entire dataset. ARF combines multiple decision trees to improve prediction accuracy. 
        \item \textbf{Learn++.NSE} \cite{Elwell_2011}: Constructs an ensemble of models trained on previous data segments and adjusts their weights based on their performance on the current data segment. 
        \item \textbf{SEGA} \cite{song_2021}: Ensembles models trained from equal-length segments of historical data and selects segments with minimum kNN-based distributional discrepancy with the current data. 
    \end{itemize}
    \item \textbf{Data-centric Method:}
    \begin{itemize}[leftmargin=*]
        \item \textbf{CVDTE} \cite{fan_2004}: This method trains a Cross-Validation Decision Tree Ensemble classifier on individual samples that do not have conflicting predictions caused by shifted decision boundaries due to concept drifts. 
    \end{itemize}
    \item \textbf{Data Subset Selection Methods:}
    \begin{itemize}[leftmargin=*]
        \item \textbf{GLISTER} \cite{killamsetty2021glistergeneralizationbaseddata}: Trains a neural network classifier based on data subset selection, ranking data subsets by gains and selecting the top-k subsets within a predefined budget.
        \item \textbf{GRAD-MATCH} \cite{killamsetty2021gradmatchgradientmatchingbased}: Trains a neural network classifier via data subset selection, simultaneously selecting data subsets and adjusting their weights to minimize gradient error.
    \end{itemize}
    \item \textbf{Quilt} \cite{kim2024quilt}: A data-centric framework designed to identify and select data segments that maximize model accuracy, utilizing gradient-based scores for efficient data segment selection. 
\end{itemize}

\subsection{Experimental Results}

For each dataset, we report accuracy, F1 score and runtime results of our method by setting the last (latest) segment as the current segment. This means that we use the most recent segment of data to evaluate how well our method performs in a real-world scenario where data is continuously evolving. We compare our method with other baseline methods across all ten datasets, as shown in Table \ref{tab:first} and \ref{tab:second}. Our results indicate that our method consistently outperforms all the baselines in terms of accuracy. This superior performance is attributed to our method's effective utilization of drifted data, which allows it to maintain high accuracy even when the data distribution changes over time. We have used the public codebase of Quilt \cite{kim2024quilt} to get the results of all baselines.

\begingroup
\footnotesize
\begin{table*}
    \centering
    \caption{Accuracy, F1 score and runtime (sec) on synthetic datasets. \% of data used indicates how much data is used compared to Quilt with 100\% data to get best performance. Best numbers are in bold. "-" indicates either memory error or time limit error.}
    \label{tab:first}
    \begin{tabular}{l|ccc|ccc|ccc|ccc|ccc|ccc}
        \toprule
        \textbf{Methods} & \multicolumn{3}{c|}{\textbf{SEA}} & \multicolumn{3}{c|}{\textbf{RandomRBF}} & \multicolumn{3}{c|}{\textbf{Sine}} & \multicolumn{3}{c|}{\textbf{Hyperplane}}  & \multicolumn{3}{c|}{\textbf{Covcon}} & \multicolumn{3}{c}{\textbf{Covcon\_5M}} \\
        \midrule
        & \textbf{Acc} & \textbf{F1} & \textbf{Time} & \textbf{Acc} & \textbf{F1} & \textbf{Time} & \textbf{Acc} & \textbf{F1} & \textbf{Time} & \textbf{Acc} & \textbf{F1} & \textbf{Time} & \textbf{Acc} & \textbf{F1} & \textbf{Time} & \textbf{Acc} & \textbf{F1} & \textbf{Time}\\
        \midrule 
        Full Data & .849 & .881 & 3.36 & .821 & .820 & 9.43 & .449 & .436 & 2.25 & .843 &.844 & 2.41 & .561 & .576 & 1.57 & .401	& .444& 	800\\
        Current segment & .864 & .888 & 0.20 & .679 & .673 & 0.56 & .899 & .898 & 0.94 & .893 & .894 & 0.46 & .941 & .540 & 2.11& .904&	.551&	30\\
        \midrule
        HAT & .825 & .862 & 1.38 & .514 & .519 & 2.35 & .293 & .305 & 1.67 & .862 & .862 & 2.23 & .83 & .674 & .733 &-&-&-\\
        ARF & .825 & .863 & 23.49 & .645 &.642 & 44.64 & .821 & .823 & 21.40 &.793 &.793& 26.86 & .873 & .89 & 9.689 &-&-&-\\
        Learn++.NSE &.804  &.836& 6.65& .611  &.612& 5.68& .925  &.925& 5.73& .755 &.755& 7.05 & .690 & .671 & 6.007  &-&-&-\\
        SEGA & .797  &.842& 4.37& .825  &.825& 4.47& .253  &.260& 4.35& .851  &.851 &4.49& .786& .601&0.67 &-&-&-\\
        \midrule
        CVDTE & .806 &.810& 0.02& .614  &.621& 0.05& .857  &.835& 0.02& .752  &.752&0.04& .940& .540&0.06 &-&-&- \\
        \midrule
        GLISTER & .857  &.885& 25.89& .794  &.794& 63.73& .879  &.876& 14.93& .905 &.905& 21.64&-&-&- &-&-&-\\
        GRAD-MATCH & .853  &.884& 2.13& .790  &.790& 6.66& .547  &.529& 0.80& .845  &.845& 1.40&-&-&-&-&-&-\\
        \midrule
        Quilt & .893  & {.909} & .993& .833  &.833& 2.836& .938  &.936& 3.909& .911  &.912& 1.810&\textbf{.988}&.918&.95& .923&	.415&	27\\
        \midrule
        \textbf{Our Method}  & \textbf{.899}  &\textbf{.912}&  1.457&  \textbf{.839}  &\textbf{.838}& 3.382&  \textbf{.955}  &\textbf{.949}& 4.369&  \textbf{.924}  &\textbf{.922}&  2.681&\textbf{.988}&\textbf{.922}&1.44 & \textbf{.968}	& \textbf{.603}	& 731\\
        {\% of data used} & \multicolumn{3}{c|}{(89.14\%)} & \multicolumn{3}{c|}{(93.43\%)} & \multicolumn{3}{c|}{(94.25\%)} & \multicolumn{3}{c|}{(87.94\%)}  & \multicolumn{3}{c|}{(62.28\%)} & \multicolumn{3}{c}{(68.49\%)}\\
        \bottomrule
    \end{tabular}
\end{table*}
\endgroup

\begingroup
\footnotesize
\begin{table*}
    \centering
    \caption{Accuracy, F1 score and runtime (sec) on real datasets. \% of data used indicates how much data is used compared to Quilt with 100\% data to get best performance. Best numbers are in bold. "-" indicates either memory error or time limit error.}
    \label{tab:second}
    \begin{tabular}{l|ccc|ccc|ccc|ccc|ccc|ccc}
        \toprule
        \textbf{Methods} & \multicolumn{3}{c|}{\textbf{Electricity}} & \multicolumn{3}{c|}{\textbf{Weather}} & \multicolumn{3}{c|}{\textbf{Spam}} & \multicolumn{3}{c|}{\textbf{Usenet1}}  & \multicolumn{3}{c}{\textbf{Usenet2}} & \multicolumn{3}{c}{\textbf{Covertype}} \\
        \midrule
        & \textbf{Acc} & \textbf{F1} & \textbf{Time} & \textbf{Acc} & \textbf{F1} & \textbf{Time} & \textbf{Acc} & \textbf{F1} & \textbf{Time} & \textbf{Acc} & \textbf{F1} & \textbf{Time} & \textbf{Acc} & \textbf{F1} & \textbf{Time} & \textbf{Acc} & \textbf{F1} & \textbf{Time}\\
        \midrule 
        Full Data & .694 &.758 & 7.42 & \textbf{.800} &.641 & 4.33 & .970 &.973 & 1.17 & .576 &.512 & 0.23 &.701 & .413 & 0.18 & .642 & .650 & 105\\
        Current segment & .709 &.756 & 0.52 & .756 &.509 & 0.26 &.955  &.963 & 0.16 &.752 &.716 & 0.16 &.745 & .613 & 0.18 & .598 & .584 & 1.76\\
        \midrule
        HAT & .691 &.743 & 6.43 & .729 & .452 & 2.10 & .888 & .847 & 25.67 & .622 & .558 & 0.87 & .730 & .472 & 0.87 & .536 & .538 & 250 \\
        ARF & .713 &.762 & 57.36 & .775 & .542 & 30.51 & .921 & .931 & 44.83 & .629 & .616 & 4.12 & .682 & .31 & 4.04 & .483 & .398 & 1843 \\
        Learn++.NSE & .698 &.734 & 17.26 & .703 & .523 & 7.86 & .928 & .942 & 3.81 & .433 & .412 & 0.35 & .637 & .251 & 0.33 &-&-&-\\
        SEGA & .637 &.697 & 10.26 & .777 & .602 & 4.11 & .858 & .851 & 6.67 & .403 & .318 & 0.84 & .630 & .207 & 0.84 & .513 & .488 & 553\\
        \midrule
        CVDTE & .689  &.736 & 0.04 & .731  &.497 & 0.03 & .917  &.918 & 0.12 & 0.718  &.624 & 0.01 &.689 &.523 &0.01 & .604 & .6 & 2.013 \\
        \midrule
        GLISTER & .698  &.741 & 77.46 & .793  &.649 & 40.79 & .971  &.974 & 14.52 &.771   &.736 & 2.07 &.744&.603&1.89 &-&-&-\\
        GRAD-MATCH & .686  &.748 & 5.97 & .795  &.622 & 3.51 & .968  &.972 & 1.13 & .630  &.591 & 0.13 &.679 &.420&0.12 &-&-&-\\
        \midrule
        Quilt & .831  &.775& 3.613& .776  &.635 & 3.284 & .988  &.976 & 2.498 & .892  &.782 & .324 &.771&.640&1.03 & .622 & .615 & 115\\
        \midrule
        \textbf{Our Method}  & \textbf{.833}  &\textbf{.825} &  2.528 &  .778  &\textbf{.696} & 1.302 & \textbf{.992 } &\textbf{.996} & .89 & \textbf{.904}  &\textbf{.841} & .119 & \textbf{.879} & \textbf{.794} & .147 &  \textbf{.689} & \textbf{.665} & 120\\
     {\% of data used} & \multicolumn{3}{c|}{(92.10\%)} & \multicolumn{3}{c|}{(56.28\%)} & \multicolumn{3}{c|}{(79.56\%)} & \multicolumn{3}{c|}{(81.48\%)}  & \multicolumn{3}{c|}{(73.33\%)} & \multicolumn{3}{c}{(53.42\%)}\\
        \bottomrule
    \end{tabular}
\end{table*}
\endgroup

In comparison, the Full Data method, which uses all available data including drifted data, does not perform as well because it is forced to incorporate data that may no longer be relevant to the current segment. On the other hand, the Current Segment method, which only uses the most recent segment of data, fails to leverage valuable historical data, leading to lower accuracy. HAT, another baseline, performs worse than our method because it adaptively learns from recent data without using previous models or historical data, limiting its ability to adapt effectively to data drift. The ensemble methods, including ARF, Learn++.NSE, and SEGA, also under perform compared toour method. ARF, for example, can lose useful previous knowledge when replacing an obsolete tree for drift adaptation, which negatively impacts its performance. Learn++.NSE and SEGA attempt to save all past models or a buffer’s worth of them and use the current data segment to create ensembles. However, these models, trained on previous data segments, struggle to fit the current data segment accurately with simple ensemble techniques. CVDTE, another baseline, performs worse than our method because it simply collects samples that do not have conflicting predictions, regardless of whether these samples actually benefit model accuracy. This method overlooks the importance and effectiveness of the gathered samples in enhancing the model's accuracy on the present data segment. Among the data subset selection methods, GLISTER's targeted sample selection demonstrates more consistent results compared to GRAD-MATCH's random batch selection. However, the substantial computational overhead of GLISTER renders it less feasible for use in real-time scenarios. In contrast,our method's approach of adaptively selecting suitable data segments with explicit model evaluations demonstrates the advantage of taking a data-centric approach. By explicitly evaluating and selecting the most relevant data segments,our method achieves better accuracy and efficiency, highlighting its effectiveness in managing data drift in ML systems.

\subsection{Ablation Study}

\begin{figure}[h]
  \includegraphics[width=0.9\columnwidth]{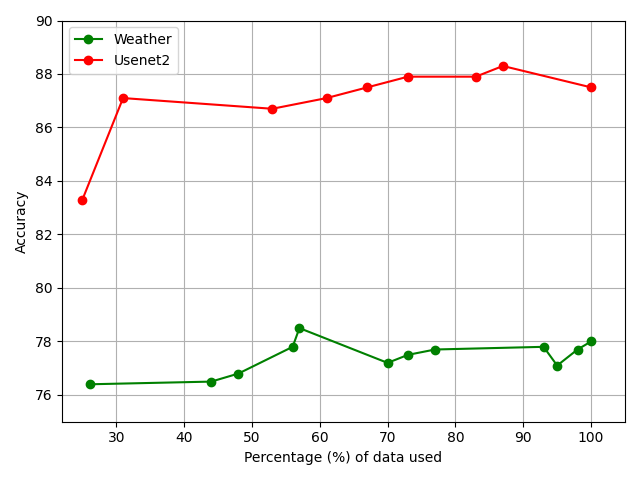}
  \caption{Trade-off between $\%$ of data used vs accuracy.}
  \Description{}
  \label{fig:mem_vs_acc}
\end{figure}

In our experiments with the Usenet2 and Weather dataset, we evaluated the effect of varying the proportion of data used for training on the model's accuracy. The goal was to determine the minimal amount of data required to achieve optimal performance. The first graph for Usenet2 (red line in Figure \ref{fig:mem_vs_acc}) presents this relationship, where we see a significant increase in accuracy when utilizing 32\% of the data, reaching approximately 87\%. This initial boost suggests that even a smaller subset of the data can capture the essential patterns necessary for effective model training. As we continue to use more data, the accuracy sees a gradual increase and then stabilizes, indicating that the additional data provides diminishing returns. The highest accuracy is observed around 87\% data utilization, after which the performance slightly decreases, reinforcing the notion that more data does not always equate to better accuracy and might even introduce noise or redundancy.

In the second graph for Weather (green line in Figure \ref{fig:mem_vs_acc}), depicting the weather dataset, the highlighted points mark a significant insight into data efficiency. At 58\% data utilization, the model reaches its peak accuracy of 78.5\%, which is higher than the accuracy obtained using the entire dataset. This indicates an optimal subset of data that maximizes the model's performance while minimizing the computational resources required. Notably, the accuracy drops when nearing 100\% data utilization, which underscores the importance of strategic data selection over sheer volume. The pattern observed here suggests that careful curation of training data segments, focusing on the most relevant subsets, can lead to superior model performance and operational efficiency. Our ablation study on both the Usenet2 and weather datasets reveals that optimal performance can be achieved with significantly less data than the full dataset. By focusing on the most relevant data, we can maintain or even improve model accuracy, making the training process more resource-efficient and effective. These findings highlight the importance of strategic data selection in developing robust and scalable ML models. The ablation of using just Algorithm \ref{alg:covariate} and training time across the modules is shown in the Table \ref{tab:runtimes}.


\begingroup
\footnotesize
\begin{table}
  \caption{For each dataset, we show the runtime results for our method, including random forest (RF) training time (Algorithm \ref{alg:covariate}), Model training time (Algorithm \ref{alg:concept}), and total time in seconds. We also provide an ablation when only Algorithm \ref{alg:covariate} is used against both algorithms are used.}
  \label{tab:runtimes}
  \begin{tabular}{l|c|c|c|c|c}
    \toprule
    \textbf{Dataset} & 
    \begin{tabular}{@{}c@{}}RF \\ train time\end{tabular}  & 
    \begin{tabular}{@{}c@{}}Model \\ train time\end{tabular}  & 
    \begin{tabular}{@{}c@{}}Total \\ train time\end{tabular}  &
    \begin{tabular}{@{}c@{}} Only Alg. \\ 1\end{tabular} &
    \begin{tabular}{@{}c@{}} Alg. \\ 1 and 2\end{tabular} \\
    \midrule
    &&&& \multicolumn{2}{c}{Accuracy} \\
    \midrule
    SEA & 1.213 & 0.244 & 1.457 & .784	& .899 \\
    Random RBF & 1.360 & 2.022 & 3.382 & .704	& .839 \\
    Sine & 1.238 & 3.131 & 4.369 & .274	& .955 \\
    Hyperplane & 1.252 & 1.429 & 2.681 & .733	& .924 \\
    Covcon & 0.710 & 0.303 & 1.441 & .421	& .988 \\
    Covcon\_5M &  707&24  & 731 & .709	& .968 \\
    \hline
    Electricity & 1.437 & 1.476 & 2.528  & .718	& .833 \\
    Weather & 0.590 & 0.712 & 1.302  & .775	& .778 \\
    Spam & 0.276 & 0.614 & 0.890  & .883	& .992 \\
    Usenet1 & 0.035 & 0.084 & 0.119  & .808	& .904 \\
    Usenet2 & 0.037 & 0.056 & 0.147  & .771	& .879 \\
    Covertype & 46 & 74 & 120 & .647	& .689 \\
    
    \bottomrule
  \end{tabular}
\end{table}
\endgroup

\section {Limitations \& Conclusion}

In this paper, we addressed the critical issue of data drift in ML systems by introducing a novel, scalable, and flexible framework. It integrates data-centric approaches with adaptive management of both covariate and concept drift. Our framework employs advanced data segmentation techniques to identify optimal data batches that reflect test data patterns, ensuring models remain relevant and accurate over time. This approach enhances model robustness by including drifted data into the training process, minimizes resource consumption, and reduces computational overhead, leading to significant cost savings. Our results on synthetic and real datasets demonstrate significant improvements in accuracy, operational cost reduction, and faster ML inference compared to state-of-the-art solutions. However, limitations such as challenges in identifying and segmenting data batches in dynamic environments and computational complexity in real-time data segmentation remain. Future work will refine segmentation techniques for better drift detection and model adaptation, conduct extensive tests on diverse datasets, and develop systems to adjust data segment and model importance based on temporal performance. Enhancements will include advanced similarity metrics using deep feature representations and temporal patterns for improved model selection. We aim to inspire further research into effective data drift solutions, enhancing the practicality and reliability of ML systems across applications.


\bibliographystyle{ACM-Reference-Format}
\bibliography{ref}


\end{document}